\pdfoutput=1

\documentclass[11pt]{article}

\usepackage[]{acl}

\usepackage{times}
\usepackage{latexsym}   

\usepackage[T1]{fontenc}

\usepackage[utf8]{inputenc}

\usepackage{microtype}

\usepackage{inconsolata}

\usepackage{subfigure}
\usepackage{array}
\usepackage{geometry}
\usepackage{booktabs}
\usepackage{amsmath}
\usepackage{graphicx}
\usepackage{hyperref}
\usepackage{url}
\usepackage{wrapfig}
\usepackage{subcaption}
\usepackage{colortbl}
\usepackage{xcolor}
\title{Unveiling Attractor Cycles in Large Language Models: A Dynamical Systems View of Successive Paraphrasing}

\author{ 
 Zhilin Wang$^{\spadesuit}\footnotemark[1]$\hspace{0.5mm},
 Yafu Li$^{\spadesuit}\footnotemark[1]\footnotemark[2]$\hspace{0.5mm},
  Jianhao Yan$^{\diamondsuit \clubsuit}$\hspace{0.5mm} \\
\bf{Yu Cheng$^{\heartsuit}$\hspace{0.5mm} \hspace{0.5mm} 
 Yue Zhang$^{\clubsuit }$}\hspace{0.2mm}\hspace{1.5mm} \\
$^\spadesuit$ Shanghai AI Laboratory\ \ \ \quad$^\clubsuit$Westlake University \\\quad$^\diamondsuit$  Zhejiang University \ \ \quad$^\heartsuit$Chinese University of Hong Kong \\
 \texttt{\{linzwcs,yafuly,elliottyan37\}@gmail.com}  \\
 \quad\texttt{\{chengyu\}@cse.cuhk.edu.hk} 
 \quad\texttt{\{zhangyue\}@westlake.edu.cn}\\
}

\begin{document}
\maketitle
\begin{abstract}
\renewcommand{\thefootnote}{\fnsymbol{footnote}}
\footnotetext[1]{\ Equal contribution. Work was done during Zhilin Wang's internship at Shanghai AI Laboratory.}
\footnotetext[2]{\ Corresponding author.}

Dynamical systems theory provides a framework for analyzing iterative processes and evolution over time. Within such systems, repetitive transformations can lead to stable configurations, known as attractors, including fixed points and limit cycles. Applying this perspective to large language models (LLMs), which iteratively map input text to output text, provides a principled approach to characterizing long-term behaviors. Successive paraphrasing serves as a compelling testbed for exploring such dynamics, as paraphrases re-express the same underlying meaning with linguistic variation. Although LLMs are expected to explore a diverse set of paraphrases in the text space, our study reveals that successive paraphrasing converges to stable periodic states, such as 2-period attractor cycles, limiting linguistic diversity. This phenomenon is attributed to the self-reinforcing nature of LLMs, as they iteratively favour and amplify certain textual forms over others. This pattern persists with increasing generation randomness or alternating prompts and LLMs. These findings underscore inherent constraints in LLM generative capability, while offering a novel dynamical systems perspective for studying their expressive potential.

\end{abstract}

\section{Introduction}

\begin{figure}[t!]
    \centering
    \includegraphics[width=0.99\linewidth]{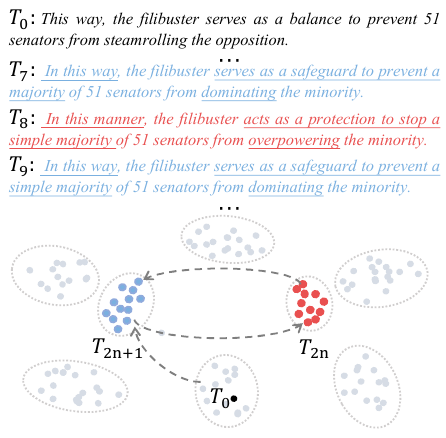}
    \caption{An illustration of successive paraphrasing using GPT-4o-mini: Here, $T_0$ denotes the original human-written text, while $T_i$ indicates the i-th round of paraphrases. The nodes depicted in the lower section represent valid paraphrases for the input sentence, with distance reflecting textual variation. Successive paraphrases generated by LLMs are confined to alternating between two limited clusters, represented as blue and orange nodes.} 
    \label{figs:intro}
\end{figure}

Dynamical systems theory provides a mathematical framework for understanding how iterative processes evolve over time~\cite{system1,system2}. 
In such systems, repetitive transformation can guide the state of the system toward stable configurations, known as attractors~\cite{attractor1}. 
These attractors can manifest as fixed points, limit cycles, or more complex structures. 
Applying this perspective to large language models, which iteratively map input text to output text, allows us to characterize their long-term behavioral patterns in a principled manner.

Paraphrase generation can serve as a valuable testbed for exploring these dynamics. 
Paraphrases are re-expressions of the same underlying meaning, differing only in their textual or linguistic form~\cite{bhagat_what_2013}. 
They serve multiple purposes: improving the readability of text for language learners~\cite{motlagh_impact_nodate,roe_what_2022,kim_how_2024}, enriching datasets in low-resource scenarios~\cite{okur_data_2022,sobrevilla_cabezudo_investigating_2024}, and enhancing stylistic variation~\cite{krishna-etal-2020-reformulating}. 
With recent advances in LLMs~\cite{touvron2023llama2openfoundation,alpaca,gpt3,gpt4}, machine-generated paraphrases can rival or surpass human quality, exhibiting remarkable generalization across diverse domains and text lengths.

While producing a single paraphrase demonstrates an LLM’s ability to exploit its prior knowledge to create textual variety while preserving semantic equivalance, \textbf{successive paraphrasing} pushes this capacity further. 
Instead of generating just one re-expression, the model recursively paraphrases its own output over multiple rounds~\cite{can_ai,ship}. 
Intuitively, this iterative process is expected to explore an expansive linguistic landscape, generating a rich tapestry of forms. Each subsequent paraphrase, based on previously transformed text, could theoretically diverge into increasingly varied structures—similar to depth-first exploration of the paraphrase search space in contrast to breadth-first approaches like beam search~\cite{Holtzman2020The,huang-etal-2023-affective,meister-etal-2023-locally}.

In practice, however, we find that this expected variety does not materialize. 
Instead of diverging across a vast combinatorial space, the LLM’s successive paraphrasing converges onto a limited set of recurring solutions, as depicted in Figure~\ref{figs:intro}. 
When studied through the lens of dynamical systems, these recurring solutions resemble \textit{a stable attractor cycle—a low-order periodic orbit in the space of possible paraphrases}~\cite{attractor1}. 
Rather than continuously discovering new linguistic configurations, the model settles into a pattern where the paraphrased outputs repeat with a fixed period. 
This phenomenon is subtle: it does not always manifest as explicit repetition but rather as a recurring rotation among a small set of structurally similar forms. 
Such periodic attractors challenge the intuition that longer or more complex texts should accomodate a broad array of distinct paraphrases. 
Instead, the LLM gravitates toward a small closed orbit, revealing inherent limitations in its expressive variability.

Specifically, to investigate this attractor-like behavior, we compile a diverse collection of human-written texts~\cite{li2023mage} and prompt a range of both open-source and commercial LLMs to perform 15 rounds of successive paraphrasing. 
Using normalized Levenshtein distance to quantify textual variation, we consistently observe a \textbf{2-period cycle}: each new paraphrase resembles the one generated two steps prior. 
This periodicity proves robust, remaining consistent across multiple models, text lengths, and prompts.
We further analyze model perplexity and generation diversity as successive paraphrasing unfolds. 
The results indicate that, rather than wandering freely in the paraphrase space, LLMs grow increasingly confident in a narrow set of solutions, effectively collapsing onto these attractors. 
Modifying generation hyperparameters or introducing perturbations, such as alternating prompts and models, only subtly disrupts these obstinate attractor cycles.
Moreover, this tendency to settle into attractor cycles extends beyond paraphrasing. Any invertible task, i.e., one that allows reconstructions of previous inputs, shows similar behavior, suggesting that such cycles are a general characteristic of LLM iterative behavior.

Finally, we propose a straightforward method to disrupt attractor cycles while maintaining semantic fidelity.
By intervening in the iterative process, we can reintroduce meaningful variation and prevent the model from settling into stable yet constrained periodic orbits.
In summary, we propose to leverage successive paraphrasing to reveal that LLM outputs, when treated as a dynamical system, tend to converge onto stable attractor cycles rather than exploring open-ended linguistic variety. 
Understanding these attractors and identifying strategies to escape them is key to unlocking the full expressive potential of LLMs.
We will release our data and code after the anonymous period.

\section{Successive Paraphrasing as System Function}



In this section, we briefly introduce the theoretical framework of dynamical systems and applies it to understand the iterative process of successive paraphrasing. 
By viewing paraphrase generation as the repeated application of a transformation (the LLM’s paraphrasing function), we connect observed phenomena, e.g., periodicity and convergence, to well-studied concepts in systems theory. 

\subsection{Systems Theory Foundations}

Systems theory provides a broad mathematical and conceptual framework for analyzing how complex processes evolve over time~\cite{system1}. 
The core idea is modeling the state of a system and its evolution through deterministic or stochastic rules. 
In continuous or discrete time, systems can exhibit distinct behaviors, ranging from stable equilibria to oscillatory dynamics or even chaotic patterns.

A \textbf{dynamical system} is commonly defined as a set of states and a rule describing how those states vary under iteration. 
When a transformation repeatedly maps an initial state to a new state, one of several outcomes often emerges:
\textit{Fixed Points}: States that remain unchanged under the transformation, representing equilibrium;
\textit{Limit Cycles}: Closed loops of states that recur periodically, representing sustained oscillations;
\textit{More Complex Attractors}: Patterns to which the system’s trajectories converge, including chaotic attractors.

These attractors shape the long-term behavior of the system. If an initial state lies within the basin of attraction of a limit cycle, for example, the system will converge to that cycle regardless of small perturbations. Identifying such attractors offers valuable insights into the stability and variability of the system’s evolution.

\subsection{Framing Successive Paraphrasing as a Dynamical System}
Successive paraphrasing involves iteratively generating variations of a given text while maintaining semantic equivalence, where each iteration builds upon the previous output. 
We propose viewing successive paraphrasing as a discrete dynamical system. 
Let $\mathcal{T}$ be the space of all possible texts. 
Consider a large language model that defines a paraphrasing function: $P: T \rightarrow T$,
where $P(T)$ outputs a paraphrase of the input text $T$. 
Given an initial text $T_0 \in \mathcal{T}$, successive paraphrasing generates the sequence $\{ T_n \}_{n=0}^\infty$ recursively by:
\begin{equation}
T_{n+1} = P(T_n), \quad n = 0, 1, 2, \dots    
\end{equation}

The set $\mathcal{P}(T)$ denotes the complete text space for valid paraphrases of $T$, which is assumed as a finite space.
In theory, the space of potential paraphrases $\mathcal{P}(T)$ can be vast, especially as text length grows. 
Each new iteration can potentially explore fresh textual variations, e.g., new syntactic structures, vocabulary choices, and stylistic nuances, while maintaining semantic equivalence.
From a systems perspective, if the mapping $P$ is capable of diversifying output states, one might expect the generated text sequence to spread broadly through the space $\mathcal{P}(T)$, never stuck in repetitive patterns, resembling a system without stable attractors. 
In contrast, if the LLM’s internal biases lead to favouring certain textual forms, the sequence may enter a basin of attraction and converge onto a stable set of states. 
In other words, rather than exhibiting limitless variety, the system might find itself drawn to limit cycles, i.e., periodic attractors in the paraphrase space.

\begin{figure*}[t!]
    \centering
    \includegraphics[width=1\textwidth]{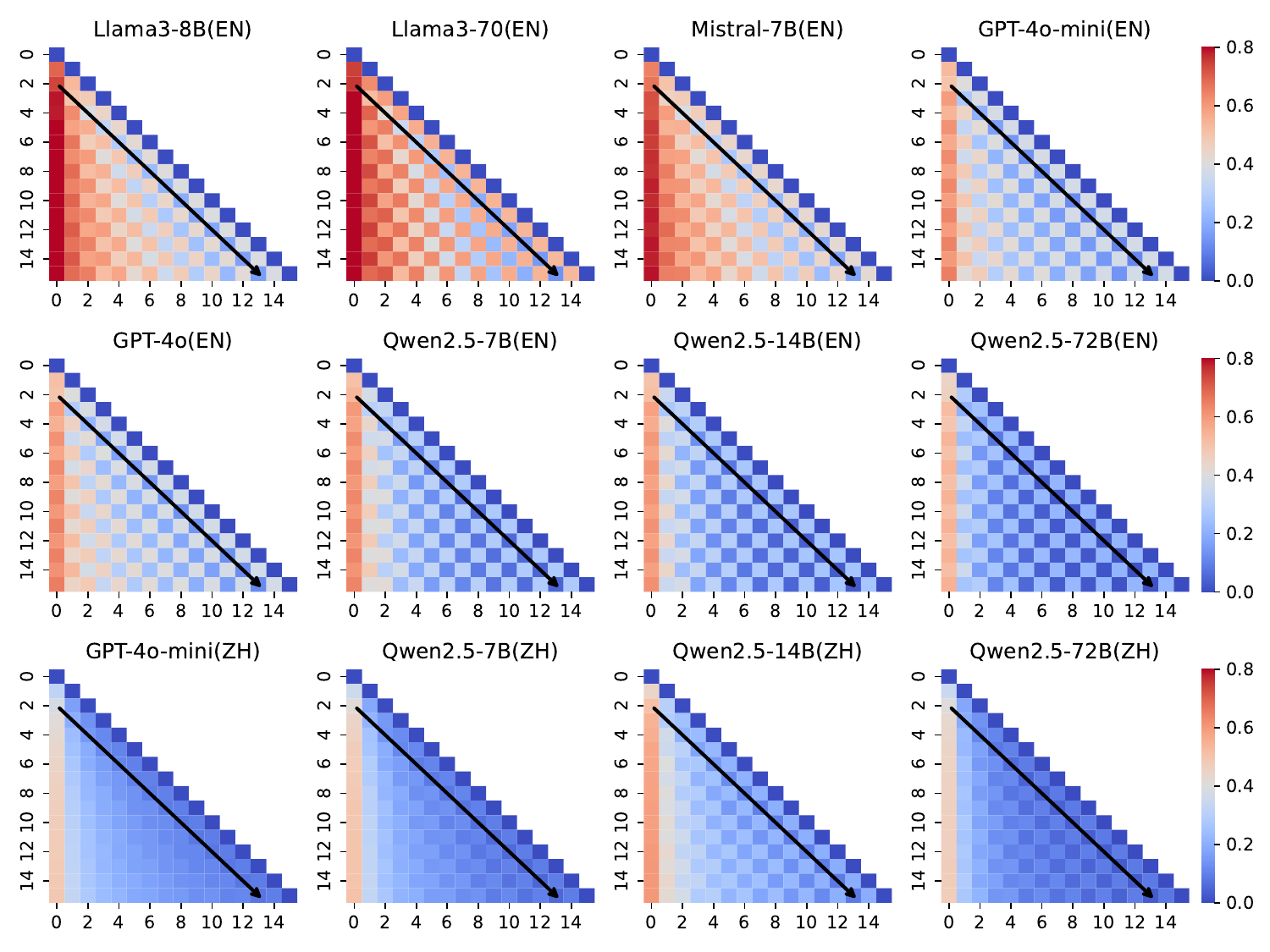}
    \caption{The difference confusion matrix for successive paraphrasing, where EN and ZH denotes English and Chinese sentence-level paraphrase generation accordingly. Both the x and y axes represent paraphrases at each step, and the value at the ($i$-th, $j$-th) grid position indicates the difference between the paraphrases at the $i$-th and $j$-th positions. 
    A darker color indicates a smaller difference value between two paraphrases. 
    The black arrow underlines the differences between \(T_i\) and \(T_{i-2}\), and averaging these values and subtracting the result from 1 gives our 2-period degree $\tau$.
    } 
    \label{figs:periodicity}
\end{figure*}

\section{Experiment Setup}
To systematically investigate this pattern, we first build dedicated testbeds and evaluation criteria.
\paragraph{Source Data Collection.}
We consider English and Chinese paraphrasing in this work.
For English paraphrase generation, we collect human-written source documents by sampling instances from the MAGE dataset~\cite{li2023mage}.
Specifically, we uniformly collect 1,000 sentences and 30 paragraphs from each domain in the dataset.
This results in a total of 1,000 sentences and 300 paragraphs for subsequent paraphrasing.
For Chinese, we source 200 sentences from WMT 2019 \citep{barrault-etal-2019-findings} and 200 sentences from Wikipedia \cite{wikidump}. 
Detailed data statistics is presented in Appendix~\ref{app:data_stat}.
The main experiments (Section~\ref{sec:main}) utilize sentence-level paraphrasing datasets, while analytic experiments employ paragraph-level datasets to demonstrate the generality of our findings (Section~\ref{sec:Analysis}).

\paragraph{Paraphrase Generation.}
For English paraphrasing, we utilize Mistral-7B-Instruct-v0.3 ~\cite{jiang2023mistral7b}, Meta-Llama-3-8B-Instruct ~\cite{touvron2023llama2openfoundation}, Meta-Llama-3-70B-Instruct~\cite{touvron2023llama2openfoundation}, Qwen2.5-7B-Instruct, Qwen2.5-14B-Instruct, Qwen2.5-72B-Instruct~\cite{qwen2}, GPT-4o-mini and GPT-4o~\cite{openai2024gpt4technicalreport}.
For Chinese, we use  Qwen2.5-7B-Instruct, Qwen2.5-14B-Instruct, Qwen2.5-72B-Instruct, and GPT-4o-mini for paraphrase generation.
By default, we set the temperature to 0.6 and p to 0.9 during the decoding process.
We sample 10 different paraphrases at each step by setting the number of search beams to 10 and sequentially rephrasing each sample for 15 rounds. 
We select the candidate with the highest probability for the next paraphrasing iteration. 

\paragraph{Evaluation Metrics.}
We use the normalized Levenshtein edit distance function $d$ to quantify the textual differences between two paraphrases.
To provide a more intuitive of the attractor cycle, we propose a metric termed 2-periodicity degree to quantify and study the cyclic pattern in successive paraphrasing.
The 2-periodicity degree \(\tau\) is defined as $\tau = 1 - \frac{1}{M-2} \sum_{i=3}^{M} d(T_{i}, T_{i-2})$, which captures the average textual similarity between the current paraphrase and that from two steps prior. 
$M$ denotes the total number of paraphrasing iterations. 
A higher $\tau$ indicates stronger periodicity, i.e., similar between two paraphrases. 
For instance, if successive paraphrases exhibit perfect 2-periodicity such that \(d(T_{i}, T_{i-2}) = 0\), then \(\tau = 1\), indicating that the current paraphrase matches exactly with that from two steps earlier. 
To evaluate semantic equivalence, we employ cosine similarity on sentence embeddings~\footnote{https://huggingface.co/sentence-transformers/all-MiniLM-L6-v2}~\cite{sentence_embed}.

\begin{table*}[t!]
    \centering
    \small
    \begin{tabular}{cccccc}
        \toprule
        \rowcolor{blue!20} \textbf{Mistral-7B} & \textbf{Llama3-8B} & \textbf{Llama3-70B} & \textbf{GPT-4o-mini} & \textbf{GPT-4o} & \textbf{Qwen2.5-7B} \\
        \midrule
        0.71 & 0.72 & 0.60 & 0.83 & 0.81 & 0.86 \\
        \midrule
        \cellcolor{blue!20} \textbf{Qwen2.5-14B} &\cellcolor{blue!20}  \textbf{Qwen2.5-72B} &  \cellcolor{red!20}  \textbf{Qwen2.5-7B} & \cellcolor{red!20}  \textbf{Qwen2.5-14B} &  \cellcolor{red!20} \textbf{Qwen2.5-72B} & \cellcolor{red!20} \textbf{ GPT-4o-mini} \\
        \midrule
        0.89 & 0.92 & 0.70 & 0.84 & 0.92 & 0.88 \\
        \bottomrule
    \end{tabular}
     \caption{The periodicity degree $\tau$ of different LLMs. The models represented in \colorbox{blue!20}{blue} denotes the English paraphrase generation, while\colorbox{red!20}{red} indicating Chinese paraphrasing.}
    \label{table:periodicity}
    \label{tab:model_comparison}
\end{table*}

\begin{figure*}[h]
    \centering
    \includegraphics[width=1\linewidth]{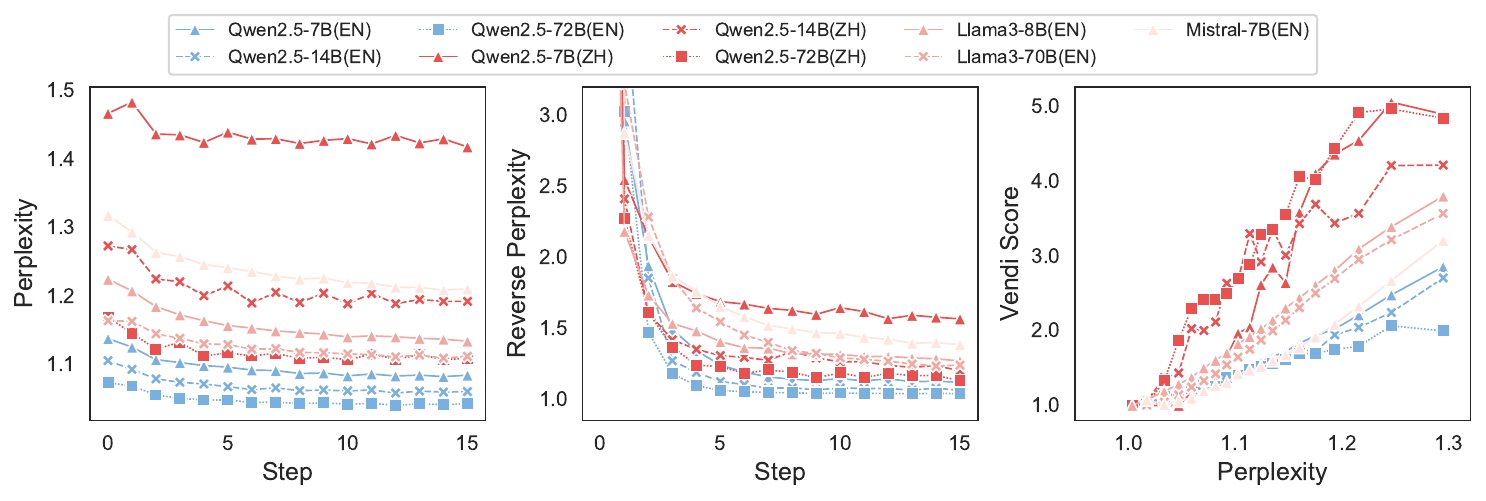}
    \caption{Convergence of perplexity, reverse perplexity, and generation diversity. The left and middle plots show that as the number of steps increases, both perplexity and reverse perplexity decrease steadily until they reach their lower bounds. The right plot shows that generation decreases as perplexity decreases.} 
    \label{figs:convergence}
\end{figure*}

\section{Results}
\label{sec:main}
Building on the dynamical systems perspective introduced earlier, we now examine the empirical evidence that successive paraphrasing leads LLMs toward stable attractor cycles. 
We iteratively paraphrase sentences over 15 rounds within the sentence-level dataset and calculate the 2-periodicity degree.

\subsection{Periodicity}
\label{Periodicity}
We calculate the textual difference between $T_i$ and $T_{i-2}$ for paraphrases at each step. 
Arranging these differences into a confusion matrix (Figure~\ref{figs:periodicity}) reveals a pronounced 2-period cycle. 
For all LLMs, the matrix’s alternating light and dark patterns indicate that paraphrases generated at even iterations cluster together, and similarly, those at odd iterations form another cluster. 
This clear partitioning aligns with the behavior of a dynamical system converging onto a 2-period limit cycle—an attractor that draws the iterative process into a stable oscillation between two distinct states.

We also quantify this periodicity across different LLMs, as shown in Table~\ref{table:periodicity}. While all models exhibit some degree of 2-periodicity, Qwen2.5-72B shows a particularly strong and consistent cycle in both English and Chinese, whereas Llama3-70B displays relatively weaker periodic behavior. 
Models with higher periodicity tend to retain more semantic fidelity, suggesting that the recurring attractor states preserve core meaning even as they oscillate between two paraphrastic forms, as shown in Appendix~\ref{app:similarity}. 

While this periodicity can be viewed as an implicit repetition issue, it differs from explicit repetition of previously seen context. 
Instead, the model implicitly cycles through a limited set of paraphrastic forms without directly referencing prior iterations. 
In terms of systems theory, the model’s mapping function $P$ creates a dynamical environment in which the state space is not fully explored, with the trajectories settling into a 2-period attractor.

\subsection{Convergence to Stable Attractor}
\label{sec:Convergence}

To probe the internal dynamics that lead to these attractor cycles, we explore generation determinism with successive paraphrasing unfolds.
We define a \textbf{conditioned perplexity} $\sigma(T_i \mid T_{i-1})$, reflecting the model’s confidence in generating $T_i$ given $T_{i-1}$, and a \textbf{reverse perplexity} $\hat{\sigma}(T_i \mid T_{i+1})$, indicating how easily $T_i$ could be reconstructed from $T_{i+1}$.

Figure~\ref{figs:convergence} demonstrates that as successive paraphrasing proceeds, both perplexity and reverse perplexity decrease. 
The forward direction (perplexity) quickly converges to a low boundary, while the reverse direction starts high, indicating that initially it is hard to ``go back'' from $T_{i+1}$ to $T_i$.
However, it drops fast as paraphrasing proceeds and aligns with the forward perplexity. 
Finally, the system evolves towards a state where generating $T_{i+1}$ from $T_i$ is nearly as deterministic and predictable as reconstructing $T_i$ from $T_{i+1}$.
This symmetry resembles a \textbf{stable attractor} in a dynamical system, where bidirectional predictability indicates that the system has ``locked in'' to a limit cycle.

We further quantify generation diversity by sampling multiple paraphrases at each iteration and computing the Vendi score~\cite{friedman2022vendi}. 
As shown in Figure~\ref{figs:convergence}, a low perplexity indicates a low generation diversity. 
A Vendi score of one indicates that all paraphrases in the beam are identical to each other.
As both forward and reverse perplexity decreases, the model consistently produces similar paraphrases, leaving minimal room for alternative textual trajectories. 
From a systems viewpoint, the collapse into low perplexity and low diversity states corresponds to the model settling into the basin of attraction of a periodic orbit. 
Once inside the basin, the model’s generative behavior becomes nearly deterministic, causing the output sequence to cycle predictably.

The notion of invertibility, where each paraphrase can be treated as a paraphrase of its own paraphrase, further explains the robustness of periodicity. 
Invertibility places constraints on the mapping function $P$, effectively enabling a bidirectional relationship between states which encourages stable cycles. 
This insight suggests that tasks with similar invertible properties, e.g., translation, can also display limit cycle behavior, a hypothesis we will explore in Section~\ref{sec:beyond paraphrasing}.

\section{Analysis}
\label{sec:Analysis}
In this section, we perform analytical experiments on paragraph-level paraphrase datasets to generalize our findings to longer texts.
We first demonstrate the extension of our findings to other task formats (Section~\ref{sec:beyond paraphrasing}). 
Then we go through a set of methods to try to escape from the attractor cycles in the remaining subsections.

\subsection{Beyond Paraphrase Generation}
\label{sec:beyond paraphrasing}



Our earlier results indicate that successive paraphrasing leads LLMs to settle into periodic attractors—specifically, 2-period limit cycles. According to the systems-theoretic perspective, such cycles should arise whenever the transformation is invertible, enabling a bidirectional mapping that makes prior states easily reproducible. To test this, we examine four additional invertible tasks at the paragraph level: polishing (Pol.), clarification (Clar.), informal-to-formal style transfer (I/F.), and forward/backward translation (Trans.). These tasks are defined in Appendix~\ref{App:tasks}.

\begin{figure}
\centering
\includegraphics[width=0.45\textwidth]{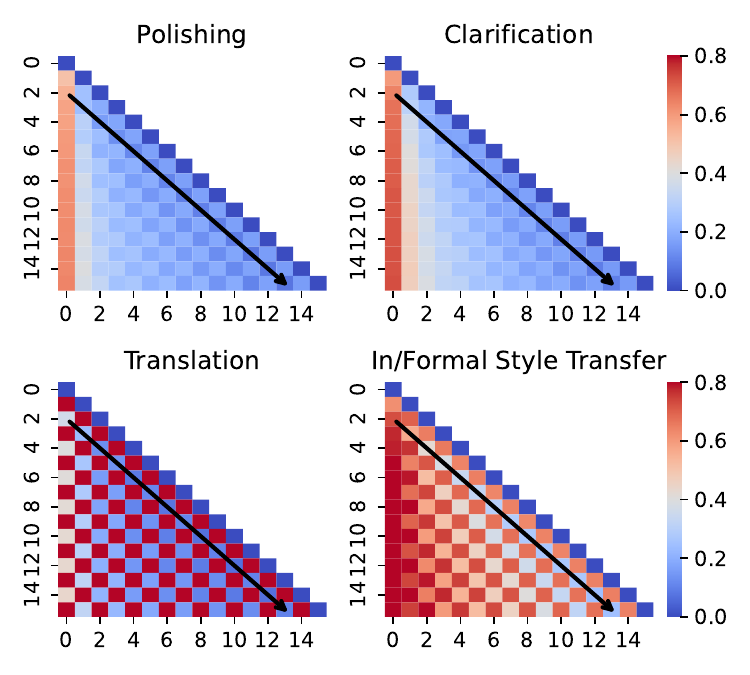}
\caption{The difference confusion matrix for four tasks beyond paraphrasing. 
Note that in translations, the difference between texts in two different languages is set to one.}
\label{figs:task_extensions}
\end{figure}

Figure~\ref{figs:task_extensions} shows that even for these varied tasks, LLMs repeatedly converge to stable states, exhibiting pronounced 2-periodicity. 
Table~\ref{table:task_extension} shows the degree of 2-periodicity across these tasks, with values ranging from 0.65 to 0.87. 
This finding reinforces the idea that invertibility fosters the emergence of limit cycles, as the model iterates the transformation and settles into an attractor. 
While paraphrasing is our primary lens, these findings confirm that stable attractor cycles are a broader characteristic of LLM behavior in iterative, invertible mappings.

\begin{table}[!h]
    \centering
    \small
    \begin{tabular}{cccccc}
        \toprule
        \textbf{Tasks}& \textbf{Para.} & \textbf{Clar.} &\textbf{Pol.} & \textbf{I/F.} & \textbf{Trans.} \\
        \midrule
        \textbf{$\tau$} &0.80 & 0.83 &  0.86 & 0.65 & 0.87 \\
        \bottomrule
    \end{tabular}
    \caption{Impact of perturbations on periodicity compared to the original during paraphrasing.}
    \label{table:task_extension}
\end{table}


%

\subsection{Alternating Models and Prompts}



\begin{figure}[!hb]
    \centering
    \begin{minipage}{0.45\textwidth}
        \centering
        \includegraphics[width=\linewidth]{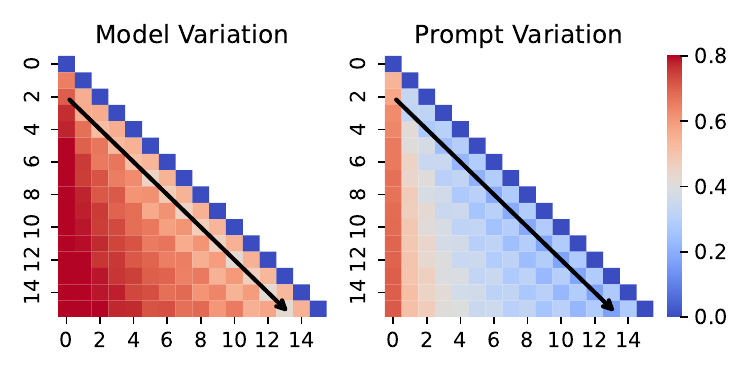}
        \caption{The difference confusion matrices for model variation and prompt variation.}
        \label{figs:model_prompt_var}
    \end{minipage}
    \hfill
    \begin{minipage}{0.45\textwidth}
          \centering
    \includegraphics[width=1\linewidth]{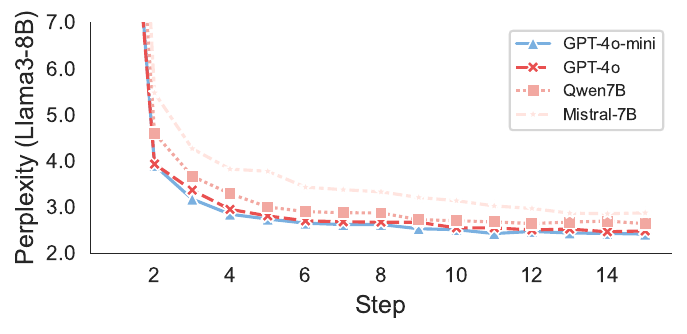}
    \caption{The perplexity of \(T_i\) conditioned on \(T_{i-1}\) calculated by Llama3-8B. Both \(T_i\) and \(T_{i-1}\) are generated by other LLMs. }    
    \label{figs:PPL_conditioned_Llama3}
    \end{minipage}
    \label{fig:overall}
\end{figure}

One intuitive approach to escape an attractor is to introduce perturbations in the transformation itself. We attempt this by varying both models and prompts during successive paraphrasing. 
For \textbf{prompt variation}, we design four different paraphrasing prompts (refer to Appendix~\ref{App:prompt_var}) and randomly select one at each iteration. Despite regularly switching prompts, the 2-period cycle persists, as shown in Figure~\ref{figs:model_prompt_var}.

Similarly, we introduce \textbf{model variation} by alternating among GPT-4o-mini, GPT-4o, Llama3-8B, and Qwen2.5-7B during successive paraphrasing. 
Although each model brings its own stylistic biases, the fundamental attractor cycle remains intact. 
Interestingly, perplexity computed by a single model (e.g., Llama3-8B) on paraphrases generated by other models still decreases over iterations in Figure~\ref{figs:PPL_conditioned_Llama3}. 
This suggests that the attractor states are not confined to a single model’s parameter space.
Instead, they reflect a more general statistical optimum that multiple LLMs gravitate toward.

From a systems perspective, this findings suggest that randomizing the transformation function $P$ does not inherently break the attractor. 
The system remains in a basin of attraction shared across these varied modeling conditions, implying that the stable cycle is a robust property of the iterative transformation rather than a quirk of any particular prompt or model.

\subsection{Increasing Generation Randomness}
\label{sec:temperature}


\begin{figure}[!h]
    \centering
    \includegraphics[width=\linewidth]{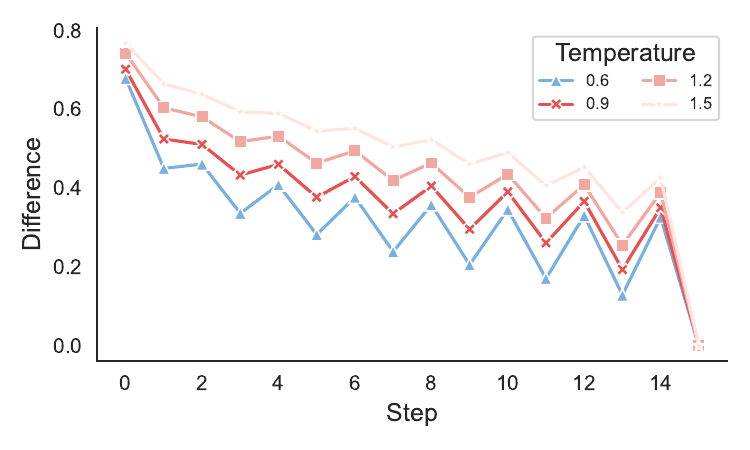}
    \caption{
    The difference between \(T_{15}\) and \(T_i\) generated by GPT-4o-mini. 
    By increasing the temperature, randomness is amplified, causing the differences to grow as well.
    }    
    \label{figs:Temperature}
\end{figure}
Another strategy is introducing more stochasticity in the generation process by increasing the generation temperature. 
Higher temperatures expand the immediate token selection space, potentially allowing trajectories to wander away from the attractor. However, as shown in Figure~\ref{figs:Temperature}, while higher temperatures do increase the difference between successive paraphrases, the system still exhibits a 2-period cycle. 
Further increases in temperature lead only to nonsensical outputs.
This outcome aligns with dynamical systems theory: a small increase in stochasticity may create local perturbations, but if the basin of attraction is strong, the system remains near the limit cycle. 
Excessive stochastic forcing can push the system out of meaningful regions of state space entirely, leading to “chaotic” or nonsensical behavior, rather than discovering a new stable attractor with richer linguistic diversity.

\subsection{Experiments with Complex Prompts}
Previous experiments were conducted using a simple paraphrase prompt, leading to existing limitations. To solve this, we experimented with a more complicated prompt, and the results indicated similar periodicity patterns. This prompt forces LLMs to enhance grammatical and syntactical variety. 
We used this prompt to instruct GPT-4o-mini to successively paraphrase the paragraph-level test set for 15 rounds. The empirical evaluation of periodicity (2-periodicity score) and convergence (PPL) of the successive paraphrasing with the complex prompt is listed below. Both the difference confusion matrix and the prompt are shown in  Appendix~\ref{app:complex_prompt}.

\begin{table}[h]
\centering
\small
\begin{tabular}{ccc}
\toprule
\textbf{Model} & \textbf{Periodicity} & \textbf{Convergence} \\ \midrule
Original & 0.80 & 1.19 \\
Complex  & 0.67 & 1.33 \\ \bottomrule
\end{tabular}
\caption{Periodicity and Convergence Table}
\end{table}

Although the sophisticated prompt alleviated the periodicity and convergence in some degree, the pattern of 2-period cycle remained strong. For context, a periodicity score of 0.67 implies an average edit distance of 0.33 between paraphrases two steps apart, whereas direct paraphrase exhibits an edit distance of 0.68.

\subsection{Incoporating Local Perturbations}

We introduce local perturbations to mitigate the attractor cycle pattern. 
At the end of each iteration, we edit 5\% of the text by introducing perturbations using three methods: synonym replacement (S.R.), word swapping (W.S.), and random insertion or deletion (I./D.).
As shown in Table~\ref{table:huamn_interven}, 
among these interventions, synonym replacement barely affects periodicity, suggesting that minor lexical changes do not move the system out of the attractor’s basin. 
It indicates that except during the first paraphrasing, LLMs primarily perform synonym replacements for words or phrases, as shown in Figure \ref{figs:intro}.
Word swapping, however, causes more significant disruption, lowering periodicity more effectively. 
From a dynamical standpoint, large structural perturbations are needed to shift the system’s state out of a stable cycle. 
Local lexical tweaks do not suffice because the attractor’s pull is strong and preserved at a deeper structural level.
\begin{table}[!h]
    \centering
    \small
    \begin{tabular}{cccc}
        \toprule
            \textbf{w/o Perturb.} & \textbf{S.R.} & \textbf{W.S.} & \textbf{I./D.} \\
        \midrule
             0.77 & 0.73 &  0.62 & 0.66 \\
        \bottomrule
    \end{tabular}
    \caption{Impact of different types of perturbations on 2-periodicity degrees $\tau$, compared to the original text during paraphrasing.}
    \label{table:huamn_interven}
\end{table}


\subsection{Paraphrasing with History Paraphrases}

\begin{figure}[!h]
    \centering
    \subfigure{\includegraphics[width=0.6\linewidth]{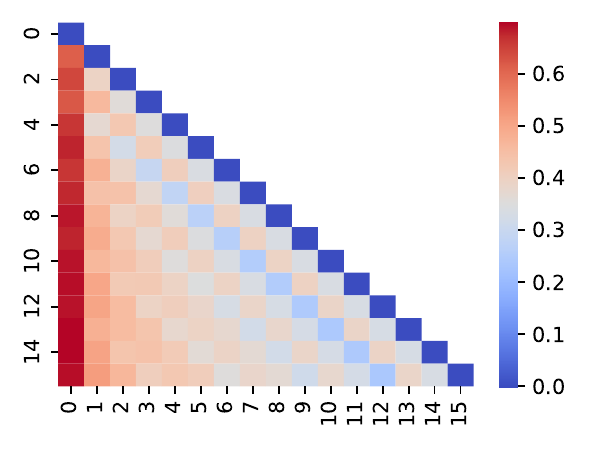}} 
    \caption{When adding historical paraphrases, LLMs exhibit 3-periodicity in the paraphrasing task.}    
    \label{figs:3-periodicity}
\end{figure}

We consider a scenario where the transformation $\hat{P}$ depends on both $T_i$ and $T_{i-1}$.
This added historical context can alter the equilibrium states. 
In a scenario where we paraphrase \(T_{i}\) based on the reference \(T_{i-1}\), it is essential that \(T_{i+1}\) differs from both \(T_{i}\) and \(T_{i-1}\). This function can be expressed as: $T_{i+1} = \hat{P}(T_{i}, T_{i-1})$.
In this context, \(P_{i-1}\) emerges as a strong candidate for paraphrasing \(P(T_{i+1}, T_{i})\), as it aligns with the distribution of LLMs while maintaining difference from \(\hat{P}(T_{i+1}, T_{i})\), satisfying the task requirement.
As a result, this more complex cycle still represents a stable attractor, albeit of higher order, as shown in Figure~\ref{figs:3-periodicity}.

\label{sec:history}


\subsection{Sample Selection Strategies}

\label{sec:sampling_strat}



\begin{figure}[h]
    \centering
    \includegraphics[width=1\linewidth]{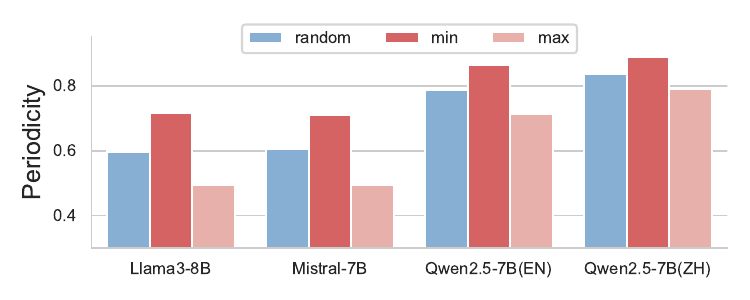}
    \caption{The periodicity of three strategies using different LLMs. } 
    \label{figs:strategy}
\end{figure}

We investigate methods to steer the system away from stable attractors at the least cost of generation quality. 
Given the correlation between periodicity and perplexity, it is intuitive to mitigate this issue by increasing perplexity while maintaining generation quality.
To achieve this, we can randomly sample multiple paraphrases at each iteration and select the one based on perplexity.
We design three types of strategies: selecting the paraphrase with the maximum or minimum perplexity or randomly choosing one at each iteration.
Figures \ref{figs:strategy} illustrate that selecting a higher perplexity can reduce periodicity.
However, such diversity comes at the cost of semantic equivalence (Appendix~\ref{app:sample_selection}).
Considering both periodicity and meaning preservation, we recommend the random strategy, which effectively reduces periodicity while incurring minimal information loss compared to selecting the option with the lowest perplexity.

\subsection{The Benefit of Mitigating Small-Size Cycles}
As a data augmentation method, paraphrase diversity should impact downstream tasks.
To figure out this, we conducted an experiment on domain classification using successive paraphrasing for data augmentation. We selected a commonly used dataset, AG News~\cite{zhang2016characterlevelconvolutionalnetworkstext} as the testbed, and trained BERT-based models using different data. For data augmentation, we conducted 5 rounds of successive paraphrasing under two different settings—min-strategy (Min.~Strat.) and max-strategy (Max.~Strat.), as detailed in Section~\ref{sec:sampling_strat}. Below are our results:

\begin{table}[h]
\centering
\small
\begin{tabular}{cccc}
\toprule
\textbf{Metric} & \textbf{w/o Aug.} & \textbf{Min. Strat.} & \textbf{ Max. Strat.} \\ \midrule
    Accuracy↑                       & 83.10\%        & 83.80\%        & 84.41\%        \\ 
2-periodicity↓    &       -     & 0.51	   &  0.33              \\ \bottomrule
\end{tabular}
\caption{Performance on AG News across different data augmentation strategies.}
\end{table}

The table shows that our max-strategy, which more effectively mitigates the 2-periodicity cycle, yields more diverse paraphrases for data augmentation and therefore achieves higher classification accuracy (84.41\%) compared to both no augmentation and the min-strategy. 

\section{Related Work}

\textbf{Paraphrase Generation.}
Paraphrase generation has long been a significant focus in NLP research, with numerous studies dedicated to enhancing the quality of generated paraphrases.~\cite{li-etal-2018-paraphrase,roy-grangier-2019-unsupervised,lewis_pre-training_2020,lin-etal-2021-towards-document-level,hosking-etal-2022-hierarchical,xie-etal-2022-multi}.
Some studies also explore methods to control paraphrase generation by focusing on aspects such as syntactic and lexical diversity
~\cite{li-etal-2019-decomposable,goyal-durrett-2020-neural,huang_generating_2021,quality,dipper,yang-etal-2022-gcpg}.
Others investigate the application of paraphrase generation as a data augmentation technique to enhance model performance~\cite{jolly-etal-2020-data,bencke2024data,okur-etal-2022-data}.

Recently, advancements in LLMs have enabled LLM-based paraphrasing tools to generate stable, high-quality responses, making them widely used for refining materials like news articles, academic papers, and speeches~\cite{witteveen-andrews-2019-paraphrasing,roe2022automated,rani-etal-2023-factify}.
However, their work primarily discusses single-step paraphrasing.
In contrast, another line of work involves LLMs iteratively rephrasing their own outputs over multiple iterations. 
\citet{can_ai} explores how repeated rephrasing can help evade AI text detectors, while \citet{ship} and \citet{huang_authorship_2024} discuss the implications for authorship after a document has undergone multiple rounds of paraphrasing.
Our research differs from those work. 
We investigate the inherent characteristics of paraphrasing when extended over multiple iterations.

\textbf{Self-Reinforcement in LLMs.}
Repetition, defined as the occurrence of repetitive text in natural language generation, has been widely explored in research community~\cite{Holtzman2020The,Welleck2020Neural,lin2021straightgradientlearninguse,see_get_2017,liu_text_2019,fu2020a}.
Several studies have observed repetition in text generation \cite{holtzman2020curiouscaseneuraltext,finlayson2023closingcuriouscaseneural} and proposed various sampling strategies to mitigate this issue. \citet{ivgi2025loopsoopsfallbackbehaviors} explores the connection between repetition and model size.
Additionally, \citet{xu_learning_2022} introduce the concept of self-reinforcement to elucidate this phenomenon, demonstrating that LLMs exhibit a tendency to repeat preceding sentences and reinforce this behavior during generation.
\citet{yan_understanding_2024} further explore the relationship between the self-reinforcement effect and the in-context learning capabilities of LLMs.
While the repetition explored in these studies primarily focuses on single-round generation scenarios, our research reveals a similar phenomenon in multi-round generation. 
We specifically examine the self-reinforcement patterns of LLMs across successive paraphrasing tasks and concentrate on typical behaviors observed in them.


\section{Conclusion}


W reframed successive paraphrasing as a discrete dynamical system, offering a principled explanation for the emergence of stable periodic attractors in LLM-generated text. 
Our empirical findings revealed that instead of producing an expanding array of diverse paraphrases, LLMs rapidly settled into low-order limit cycles. 
These attractor states persisted even when we vary models, prompts, generation temperatures, and local kicak perturbations, indicating that they stem from a fundamental property of the system rather than superficial repetition or particular model idiosyncrasies.
Viewing iterative text generation through the lens of systems theory helps clarify why certain interventions fail to break these cycles and how others can weaken the attractor’s pull. 
Ultimately, recognizing and addressing these stable attractor cycles is crucial for unlocking more expressive and flexible language generation for large language models.
%

\section*{Limitations}

While this study provides valuable insights into successive paraphrasing, several limitations should be acknowledged. 
First, the paraphrasing is based on simple prompts, which may limit the generalizability of the findings to more complex or specific prompts. Second, although we have examined this phenomenon in the currently prevalent LLMs, other LLMs may not exhibit the same behavior. 
Finally, while we present the convergence of reverse perplexity in this work, the underlying reasons for this behavior still require further investigation.

\section*{Ethic Considerations}

We uphold the Code of Ethics and ensure that no private or non-public information is used in this work. We comply with the terms set by companies offering commercial LLM APIs and extend our gratitude to all collaborators for their invaluable support in utilizing these APIs.

\section*{Acknowledgement}
We extend our gratitude to all the reviewers for their valuable feedback and suggestions, which greatly contributed to enhancing the quality of the paper. This research was supported by the Natural Science Foundation of China Key
Program under Grant Number 2022YFE0204900.




\bibliography{latex/acl_latex}

\appendix
\label{sec:appendix}

\section{Data Statistics}
\label{app:data_stat}
\begin{figure}[!h]

\centering
\includegraphics[width=0.45\textwidth]{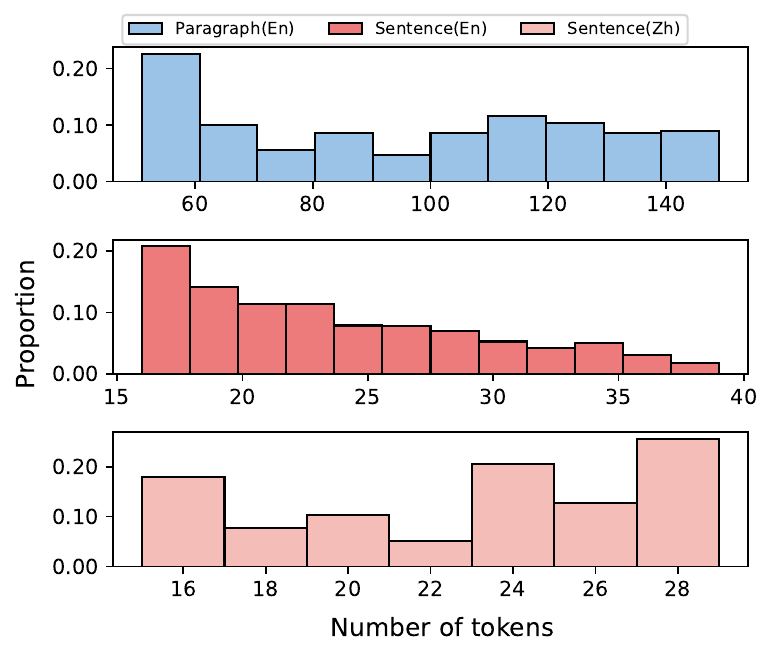}
\caption{Statistical patterns of data length distribution.}
\label{figs:data distribution}
\end{figure}

We provide source information of our data in table \ref{table:dataset} and statistic information of data length in \ref{figs:data distribution}.
\begin{table*}[!h]
    \centering
    \begin{tabular}{lcccccc}
        \toprule
        \textbf{Dataset} & \textbf{TLDR} & \textbf{SQuAD} & \textbf{ROCT} & \textbf{Yelp} & \textbf{ELI5} & \textbf{Sci\_Gen} \\
        \midrule
        \textbf{Sentence/Paragraph} & 100/30 & 100/30 & 100/30 & 100/30 & 100/30 & 100/30 \\
        \midrule
        \textbf{Dataset} & \textbf{XSum} & \textbf{CMV} & \textbf{HSWAG} & \textbf{WP} & \textcolor{red}{\textbf{Wiki}} & \textcolor{red}{\textbf{WMT}} \\
        \midrule
        \textbf{Sentence/Paragraph} & 100/30 & 100/30 & 100/30 & 100/30 & 200/0 & 200/0 \\
        \bottomrule
    \end{tabular}
    \caption{Dataset Setup: Datasets marked in red indicate Chinese datasets, while others represent English datasets. The value indicates the number of extracted samples. For example, we extract 100 sentences and 30 paragraphs from the TLDR dataset.}
    \label{table:dataset}
\end{table*}

\section{Change in similarity}
\label{app:similarity}
\begin{figure}[!h]
\centering
\includegraphics[width=0.5\textwidth]{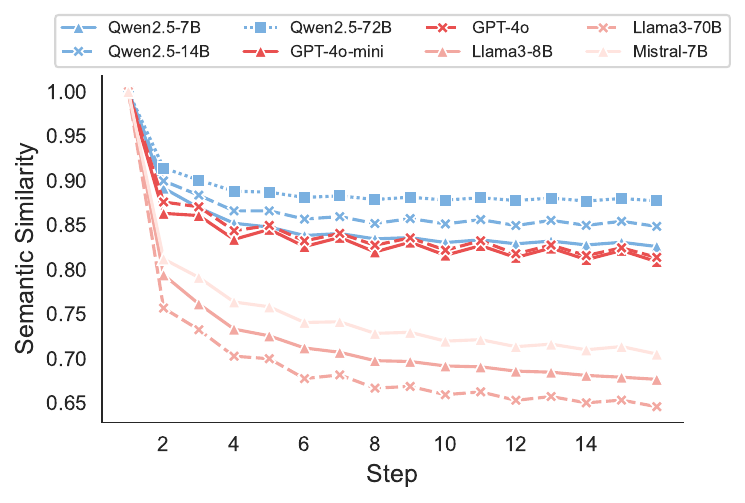}
\caption{Similarity changes during successive paraphrasing. Qwen2.5-72B is the best at preserving meaning, while all other LLMs experience slight degradation in similarity, except during the first paraphrasing step.}
\label{figs:similarity}
\end{figure}

We measure the change in similarity between \(T_i\) and \(T_0\) across successive paraphrasing steps. The results are presented in Figure \ref{figs:similarity}.
As the number of paraphrasing steps increases, most LLMs maintain the similarity between paraphrases and their corresponding original texts, with the exception of an initial drop in similarity.
Meanwhile, it also exhibits aslight 2-periodicity in similarity.
By combining Figure \ref{figs:similarity} and Table \ref{figs:periodicity}, we found that models with higher periodicity also exhibit higher similarity.

\label{App:periodicity}

\section{Generalization}
\label{app:generalization}
\label{App:Temperature}

\subsection{Task Extentions}
\begin{figure}[h]
    \centering
    \includegraphics[width=1\linewidth]{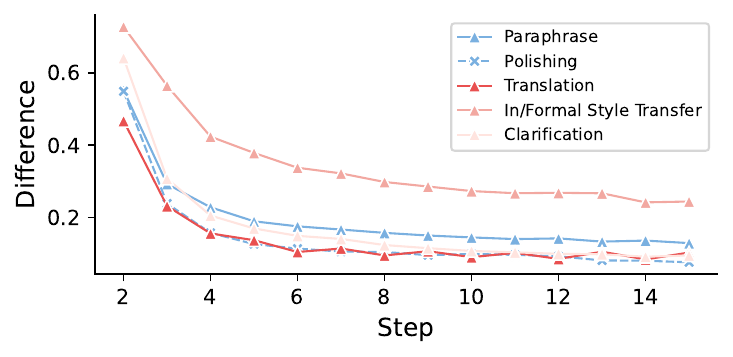}
    \caption{The trend in normalized edit distance between $T_i$ and $T_{i-2}$ across various tasks during the repetition process using GPT-4o-mini.}    
    \label{figs:Other_tasks_div_trend}
\end{figure}

We propose four additional tasks beyond paraphrasing: polishing (\textbf{Pol.}), clarification (\textbf{Clar.}), informal-to-formal style transfer (\textbf{I/F.}), and forward/backward translation (\textbf{Trans.}).
The detailed prompts for these tasks are listed in Table \ref{tab:task_extension_prompts}. 
We perform these tasks on our paragraph dataset, calculate the textual difference of the paraphrase at each iteration with the initial text, and plot the results in Figure~\ref{figs:Other_tasks_div_trend}.
As the number of paraphrasing steps increases, the difference between \(T_i\) and \(T_{i-2}\) decreases. 
After 7 steps, there is little difference between \(T_i\) and \(T_{i-2}\).


\label{App:tasks}
\begin{table}[!h]
    \centering
    \small
    \begin{tabular}{p{0.12\linewidth}p{0.78\linewidth}}
    \toprule
        \textbf{Pol.} &  Please polish the following text: \{text\}  \\
    \midrule
        \textbf{Clar.} &  Please rewrite the following text in a way that is simpler and easier to understand, using clear language and shorter sentences without losing the original meaning: \{text\}  \\
    \midrule
        \textbf{I/F.} & Transform the following text into an informal style:  \{text\} / Rewrite the following text in a formal style:  \{text\}  \\
    \midrule
        \textbf{Trans.} & Please translate the following English text into Chinese: \{text\} / Please translate the following English text into Chinese: \{text\}  \\
    \bottomrule
    \end{tabular}
    \caption{Four types of prompts for extension tasks. The last two tasks involve switching between different languages and styles, separated by a semicolon.}
    \label{tab:task_extension_prompts}
\end{table}

\subsection{Model and Prompt variation}

\label{App:prompt_var}

We continue to modify the models and prompts during paraphrasing.
The chosen model set includes GPT-4o-mini, GPT-4o, Qwen2.5-7B, and Llama3-8B.
Four variations of the paraphrasing prompts are provided in Table \ref{tab:prompts_var}.

\begin{table}[!h]
    \centering
    \small
    \begin{tabular}{p{0.05\linewidth}p{0.88\linewidth}}
    \toprule
        \textbf{A:} &  Please paraphrase the following text: \{text\}  \\
    \midrule
        \textbf{B:} &  Please rephrase the text below: \{text\}  \\
    \midrule
        \textbf{C:} & Please rewrite the following text:  \{text\}  \\
    \midrule
        \textbf{D:} & Please polish the text below: \{text\}  \\
    \bottomrule
    \end{tabular}
    \caption{Four variations of paraphrasing prompts. In the prompt variation experiments, a prompt is randomly selected at each step to perform the paraphrasing.}
    \label{tab:prompts_var}
\end{table}

\subsection{Experiment with A Complex Prompt}
\label{app:complex_prompt}
\begin{figure}
    \centering
    \includegraphics[width=0.75\linewidth]{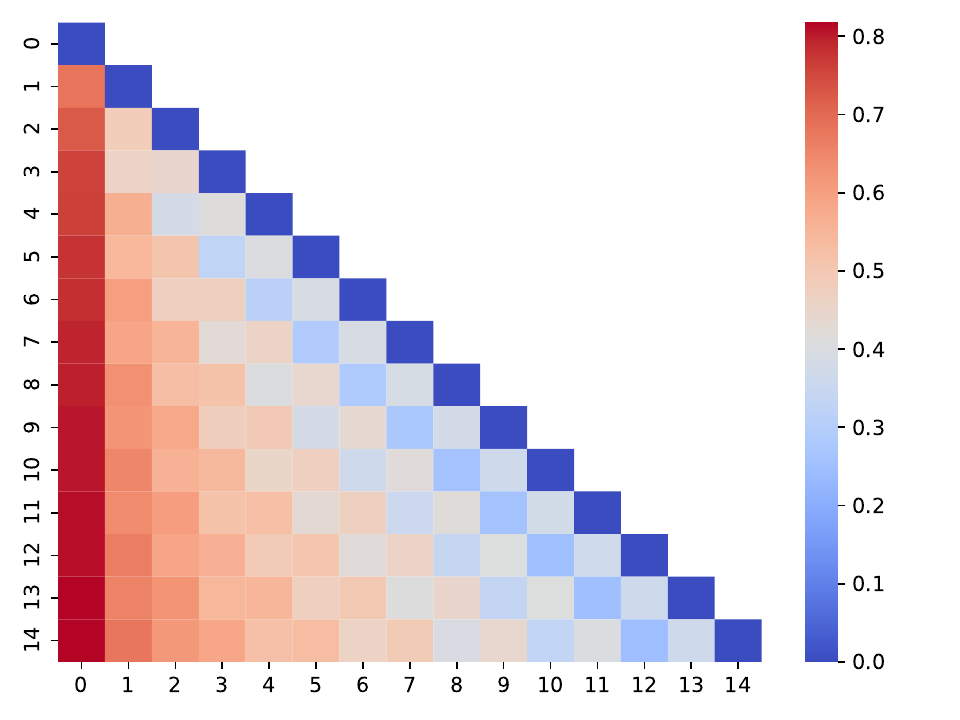}
    \caption{Difference confusion matrix for the complex prompt.}
    \label{fig:more_complex_prompt}
\end{figure}

We conduct experiments on our paragraph-level dataset using a more complex prompt, as presented in Table~\ref{tab:more_complex_prompt}. The resulting difference confusion matrix is shown in Figure~\ref{fig:more_complex_prompt}.

\begin{table}[!h]
    \centering
    \small
    \begin{tabular}{p{0.95\linewidth}}
    \toprule
    Please rewrite the following paragraph with the goal of enhancing lexical and syntactical variety without changing the original meaning. Pay attention to employing diverse vocabulary, increasing the complexity and variation of sentence structures, using different conjunctions and clause constructions to make the expression more diverse and rich, while maintaining the core information and logical coherence of the original text. Specifically, avoid repetitive sentence patterns and try to express the same ideas in different ways.\\
    \bottomrule
    \end{tabular}
    \caption{A more complex prompt enhancing lexical and syntactical variety.}
    \label{tab:more_complex_prompt}
\end{table}

\subsection{Increasing Randomness}

\begin{figure}[t!]
\centering
\includegraphics[width=0.45\textwidth]{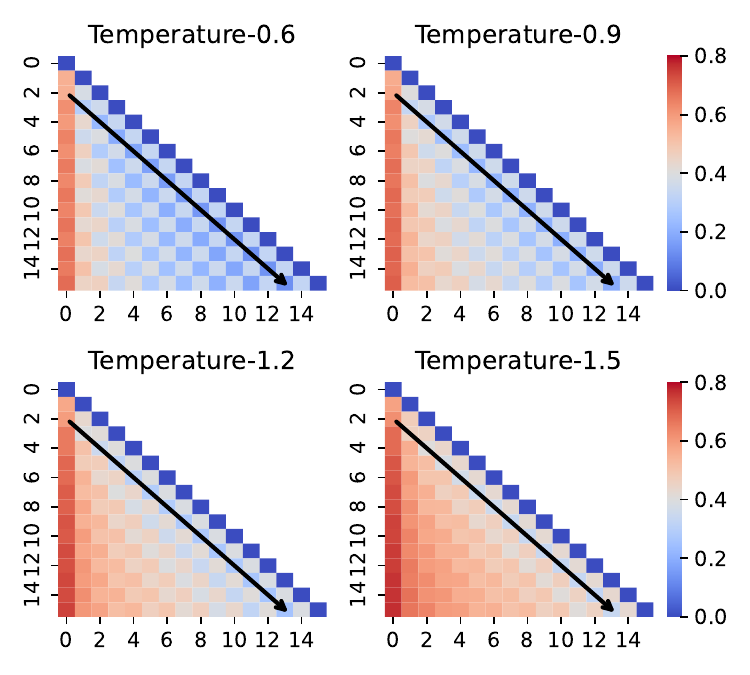}
\caption{The difference confusion matrix for successive paraphrasing at different temperature settings, conducted by GPT-4o-mini.}
\label{figs:app_temperature}
\end{figure}

We measure the impact of increasing randomness on periodicity by adjusting the generation temperature. We select four temperature values: 0.6, 0.9, 1.2, and 1.5. 
The results are shown in Figure \ref{figs:app_temperature}.
Although the temperature increases to a very high level, the 2-periodicity still persists.
Further increasing temperature will cause nonsense responses.

\begin{figure}[t!]
    \centering
    \begin{minipage}{0.45\textwidth}
        \centering
        \includegraphics[width=\linewidth]{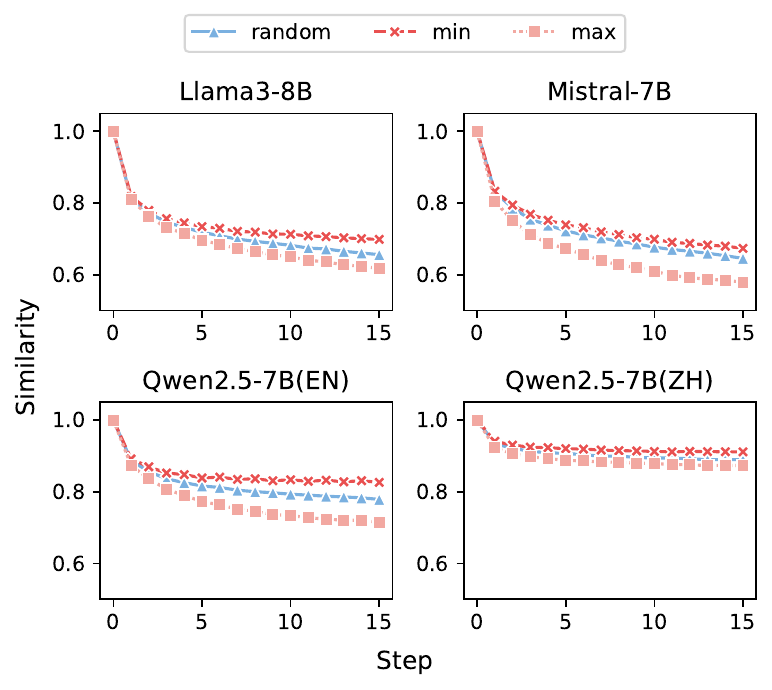}
        \caption{The similarity between paraphrases and the original texts increases during the paraphrasing process.}
        \label{figs:sim}
    \end{minipage}
    \label{fig:overall}
\end{figure}

\subsection{Sample Selection Strategies}
\label{app:sample_selection}
We propose three strategies for successive paraphrasing and evaluate them across different LLMs. 
To assess the impact of these strategies on meaning preservation, we measure the similarity between the paraphrases \(T_i\) and their corresponding original texts \(T_0\) and demonstrate the result in Figure \ref{figs:sim}.
By combining Figure \ref{figs:sim} and Figure \ref{figs:strategy}, we suggest that the random strategy preserves meaning significantly better than the max strategy, while also effectively alleviating periodicity.

\section{Case Study}
We present part of the first case from our paragraph dataset for successive paraphrasing in Figure \ref{tab:case_study_1}.


\begin{table*}[t!]
    \setlength{\belowcaptionskip}{-0.cm}
    \centering
    \small
    \renewcommand{\arraystretch}{1.2} 
    \begin{tabular}{lp{0.8\linewidth}} 
    \toprule
    \multicolumn{2}{c}{\textbf{Sample of successive paraphrasing}} \\
    \midrule
      $T_0$ &Australia has a strong tradition of an honest days pay for an honest days work. We have a strong union presence which leads to high wages backed by a strong resource economy. We also have a very multicultural society with many immigrants choosing to settle in Australia this leads to diverse shopping areas where goods originally made all over the world are made right here on our shores. The main reason Australia is the best country in the world is our heavily ingrained concept of a "fair go" meaning you give anybody a chance to prove themselves no matter who they are. This is not even mentioning our spectacular landscapes and beaches. All in all Australia is the best country in the world.  \\
    \midrule
      $ T_1$ & Australia has a robust tradition of fair compensation for hard work. The strong presence of unions contributes to high wages, supported by a solid resource-based economy. Additionally, our multicultural society, enriched by many immigrants, creates diverse shopping areas where products from around the globe are available right here in Australia. A key reason why Australia stands out as the best country in the world is our deep-rooted belief in a "fair go," which emphasizes giving everyone the opportunity to demonstrate their abilities, regardless of their background. Not to mention, our stunning landscapes and beautiful beaches add to our appeal. Overall, Australia truly is the best country in the world. \\
      \midrule
     $ T_2$ & Australia has a strong tradition of ensuring fair pay for hard work. The significant influence of unions leads to high wages, bolstered by a stable resource-driven economy. Furthermore, our multicultural society, enriched by a diverse population of immigrants, fosters vibrant shopping districts where global products are readily available. One of the main reasons Australia is considered the best country in the world is our fundamental belief in a "fair go," which highlights the importance of providing everyone with the chance to showcase their talents, regardless of their background. Additionally, our breathtaking landscapes and beautiful beaches enhance our country's appeal. Overall, Australia is truly the best place in the world.  \\
    \midrule
    $ T_3$ & Australia has a strong history of promoting fair compensation for hard work. The powerful presence of unions contributes to high wages, supported by a stable economy driven by natural resources. In addition, our multicultural society, enriched by a diverse immigrant population, creates lively shopping areas filled with international products. A key reason why Australia is often regarded as the best country in the world is our core belief in a "fair go," which emphasizes the importance of giving everyone the opportunity to demonstrate their abilities, regardless of their background. Moreover, our stunning landscapes and beautiful beaches add to the allure of our nation. Overall, Australia is undoubtedly one of the finest places in the world.  \\
    \midrule
   $T_4$ & Australia has a rich tradition of advocating for fair pay for hard work. The strong influence of unions helps maintain high wages, bolstered by a robust economy fueled by natural resources. Additionally, our multicultural society, enhanced by a varied immigrant community, fosters vibrant shopping districts filled with international goods. A significant reason Australia is often viewed as one of the best countries globally is our fundamental belief in a "fair go," which highlights the necessity of providing everyone with the chance to showcase their talents, regardless of their background. Furthermore, our breathtaking landscapes and beautiful beaches contribute to the appeal of our nation. Overall, Australia is truly one of the most exceptional places in the world.  \\
     \midrule
    $T_5$ & Australia has a rich history of promoting fair compensation for hard work. The strong presence of unions plays a crucial role in ensuring high wages, supported by a thriving economy driven by natural resources. Moreover, our diverse society, enriched by a wide-ranging immigrant population, creates lively shopping areas filled with products from around the globe. A key reason Australia is often regarded as one of the best countries in the world is our core belief in a "fair go," which emphasizes the importance of giving everyone the opportunity to demonstrate their abilities, regardless of their background. Additionally, our stunning landscapes and beautiful beaches enhance the allure of our nation. In summary, Australia is truly one of the most remarkable places in the world. \\
     \midrule
    $T_6$ & Australia has a notable history of advocating for fair pay for hard work. The significant influence of unions is essential in maintaining high wages, bolstered by a robust economy fueled by natural resources. Furthermore, our diverse society, enriched by a broad immigrant community, fosters vibrant shopping districts filled with international products. A major reason Australia is frequently seen as one of the best countries globally is our fundamental belief in a "fair go," which underscores the importance of providing everyone the chance to showcase their talents, irrespective of their background. Additionally, our breathtaking landscapes and beautiful beaches add to the appeal of our nation. In conclusion, Australia is indeed one of the most extraordinary places in the world.  \\
     \midrule
    $ T_7$ & Australia has a rich history of promoting fair compensation for hard work. The strong presence of unions plays a crucial role in sustaining high wages, supported by a thriving economy that benefits from its natural resources. Moreover, our multicultural society, enhanced by a diverse immigrant population, creates lively shopping areas filled with global products. A key reason Australia is often regarded as one of the best countries in the world is our core belief in a "fair go," which highlights the importance of giving everyone the opportunity to demonstrate their abilities, regardless of their background. Additionally, our stunning landscapes and picturesque beaches contribute to the allure of our nation. In summary, Australia is truly one of the most remarkable places on Earth.  \\
    \bottomrule
    \end{tabular}
    \caption{
    \label{tab:case_study_1}
    Case illustration of the first sample in our paragraph dataset. Due to the limited content, we just display \( T_0... T_{7}\).
    }

\end{table*}
\end{document}